\documentclass[runningheads]{llncs}
\usepackage{amsmath}
\usepackage{amssymb}
\usepackage[noend]{algpseudocode}
\usepackage{graphicx}
\usepackage{multirow}
\usepackage{booktabs,caption,fixltx2e}
\usepackage{algorithm} 
\usepackage{color}

\usepackage{pgfplots} 
\pgfplotsset{compat=1.5}

\usepackage{forest,adjustbox} 
\forestset{qtree/.style={for tree={parent anchor=south, child anchor=north,align=center,inner sep = 0pt}}}

\usetikzlibrary{decorations.markings} 
\newsavebox\mybox
\savebox\mybox{%
  \tikz{
    \draw[ultra thick,red] (-4pt,-4pt) -- (4pt,4pt);
    \draw[ultra thick,red] (-4pt,4pt) -- (4pt,-4pt);
  }%
} 
\tikzset{
    myedge/.style={
    decoration={
    markings,
    mark=at position 0.5 with \node {\usebox\mybox};
  },
  postaction=decorate
  }
}

\usepackage{threeparttable} 

\usepackage{arydshln} 
\usepackage{tikz}
\usetikzlibrary{datavisualization}
\usetikzlibrary{datavisualization.formats.functions}

\newcommand{\sd}{SSDPS}

\begin{document}
 
\title{Statistically Significant Discriminative Patterns Searching}

\author{Hoang Son Pham\inst{1} \and 
Gwendal Virlet\inst{2} \and
Dominique Lavenier\inst{2} \and
Alexandre Termier\inst{2}
}
\authorrunning{H.S. Pham et al.}

\institute{
ICTEAM, UCLouvain, Belgium \and 
Univ Rennes, Inria, CNRS, IRISA
}

\maketitle

\begin{abstract}
Discriminative pattern mining is an essential
task of data mining. This task aims to discover
patterns which occur more frequently in a class
than other classes in a class-labeled dataset.
This type of patterns is valuable in various
domains such as bioinformatics, data classification.
In this paper, we propose a novel algorithm, named \sd,
to discover patterns in two-class datasets.
The \sd\ algorithm owes its efficiency to an original enumeration strategy of the patterns,
which allows to exploit some degrees of anti-monotonicity on the measures
of discriminance and statistical significance.
Experimental results demonstrate
that the performance of the \sd\ algorithm is 
better than others. In addition, the number of generated patterns is much less than the number of the other algorithms. Experiment on real data also shows that \sd\ efficiently detects multiple
SNPs combinations in genetic data.
\end{abstract}

\begin{keywords}
Discriminative patterns, Discriminative Measures, Statistical Significance, Anti-Monotonicity.
\end{keywords}

\section{Introduction}
\label{sec:introduction}
Recently, the use of discriminative pattern mining
(also known under other terms such as emerging pattern mining
\cite{Dong1999}, contrast set mining \cite{Bay2001}) has been investigated to tackle various applications
such as bioinformatics \cite{Cheng2008}, data classification \cite{Garcia-Borroto2014}. Discriminative
pattern mining aims at finding the patterns which occur more
frequently in a class than in another class from a two-class dataset.
Various algorithms and software frameworks have been proposed
for efficiently discovering such patterns.
These algorithms can be classified into several
categories, depending on the search method and the target objective,
such as exhaustive search \cite{Li2007,TiasGuns_2011,Fang2012}, 
searching $k$ patterns which have strongest discriminative power \cite{Cong2005,Guns2013,Terlecki2008,Leeuwen2012}
and using heuristic strategies to report a good enough but not necessarily optimal result \cite{Boley2011,Garcia-Borroto2010}.
In general, most of these approaches are effective in searching for
discriminative patterns. However, some major problems remain such as
being highly time-consuming and generating a large number of patterns.

In this paper, we propose an algorithm, named \sd, that discovers
discriminative patterns in two-class datasets. More precisely, the
\sd\ algorithm aims at searching patterns satisfying both
discriminative scores and confidence
intervals thresholds. These patterns are defined as
\textbf{statistically significant discriminative patterns}.
The \sd\ algorithm is based on an enumeration strategy in which discriminative measures
and confidence intervals can be used as anti-monotonicity properties.
These properties allow the search space to be pruned efficiently.
All patterns are directly tested for discriminative scores and confidence
intervals thresholds in the mining process. Only patterns satisfying
both of thresholds are considered as the target output.
According to our knowledge, there doesn't exist any algorithms that combine
discriminative measures and statistical significance as anti-monotonicity
to evaluate and prune the discriminative patterns.

The \sd\ algorithm has been used to conduct various experiments
on both synthetic and real genomic data.
As a result, the \sd\ algorithm effectively deploys the
anti-monotonic properties to prune the search space.
In comparison with other well-known algorithms such as SFP-GROWTH
\cite{Ma2010} or CIMCP \cite{TiasGuns_2011}, the \sd\ obtains a better
performance. In addition the proportion of generated patterns is
much smaller than the amount of patterns output by these algorithms.

The rest of this paper is organized as follows:
Section \ref{sec:defs} precisely defines the concept of
statistically significant discriminative pattern, and Section \ref{sec:enum}
presents the enumeration strategy used by the \sd\ algorithm.
In Section \ref{sec:algo}, the design of the
\sd\ algorithm is described. Section \ref{sec:expes} is dedicated to
experiments and results. Section \ref{sec:conc} concludes the paper.

\section{Problem Definition}
\label{sec:defs}
The purpose of discriminative pattern mining algorithms is
to find groups of items satisfying some thresholds.
The formal presentation of this problem is given in the following:

Let $I$ be a set of $m$ items $I=\{i_1,...,i_m\}$ and $S_1$, $S_2$ be two labels. 
A $transaction$ over $I$ is a pair $t_i = \{(x_i,y_i)\}$, where $x_i \subseteq I$, $y_i \in \{S_1,S_2\}$. 
Each transaction $t_i$ is identified by an integer $i$, denoted $tid$ ({\em transaction identifier}). 
A set of transactions $T = \{t_1,..., t_n \}$ over $I$ can be termed as a {\em transaction dataset} $D$ over $I$. 
$T$ can be partitioned along labels $S_1$ and $S_2$ into $D_1 = \{t_i \; | \; t_i = (x_i, S_1) \in T\}$ and $D_2  = \{t_i \; | \; t_i = (x_i, S_2) \in T\}$. 
The associated tids are denoted $D_1.tids$ and $D_2.tids$.

For example, Table \ref{Table:exData} presents a dataset including 9
transactions (identified by $1..9$) which are described by 10 items
(denoted by $a..j$). The dataset is partitioned into two classes (class label 1 or 0).

\begin{table}
\centering
\caption{Two-class data example}
\label{Table:exData}

\begin{tabular}{| c | p{0.4cm} | p{0.4cm} | p{0.4cm} | p{0.4cm} | p{0.4cm} | p{0.4cm} | p{0.4cm} | p{0.4cm} | p{0.4cm} | p{0.4cm} | c |}
\hline
\textbf{Tids} & \multicolumn{10}{| c |}{\textbf{Items}} & \textbf{Class} \\

\hline
1 & a & b & c &   &   & f &   &   & i & j & 1 \\
2 & a & b & c &   & e &   & g &   & i &   & 1 \\
3 & a & b & c &   &   & f &   & h &   & j & 1 \\
4 &   & b &   & d & e &   & g &   & i & j & 1 \\
5 &   &   &   & d &   & f & g & h & i & j & 1 \\
\hline
6 &   & b & c &   & e &   & g & h &   & j & 0 \\
7 & a & b & c &   &   & f & g & h &   &   & 0 \\
8 &   & b & c & d & e &   &   & h & i &   & 0 \\
9 & a &   &   & d & e &   & g & h &   & j & 0 \\
\hline
\end{tabular}
\end{table}

A set $p \subseteq I$ is called an $itemset$ (or pattern) and a set $q \subseteq \{1..n\}$
is called a $tidset$. For convenience we write a tidset $\{1,2,3\}$ as $123$,
and an itemset $\{a,b,c\}$ as $abc$. The number of transactions in $D_i$ containing $p$
is denoted by $|D_i(p)|$. The {\em relational support} of pattern $p$ in class $D_i$,
denoted $sup(p,D_i)$, is defined as: 

\begin{equation}
sup(p,D_i)=\frac{|D_i(p)|}{|D_i|}
\end{equation}

The {\em negative support} of $p$ in $D_i$, denoted $\overline{\rm sup}(p,D_i)$, is defined as:

\begin{equation}
\overline{\rm sup}(p,D_i) = 1-sup(p,D_i)
\end{equation}


To evaluate the discriminative score of pattern $p$ in a two-class dataset $D$,
different measures are defined over the relational supports of $p$.
The most popular discriminative measures are
\textit{support difference}, \textit{grown rate support} and \textit{odds ratio support}
which are calculated by formulas \ref{formula:sd}, \ref{formula:gr}, \ref{formula:ors} respectively.

\begin{equation}
SD(p,D) = sup(p,D_1) - sup(p,D_2)
\label{formula:sd}
\end{equation}

\begin{equation}
GR(p,D) = \frac{sup(p,D_1)}{sup(p,D_2)}
\label{formula:gr}
\end{equation}

\begin{equation}
ORS(p,D) =\frac{ sup(p,D_1) / \overline{\rm sup}(p,D_1)} { sup(p,D_2) / \overline{\rm sup}(p,D_2)}
\label{formula:ors}
\end{equation}

A pattern $p$ is discriminative if its score is not less than a given threshold $\alpha$.
For example, let $\alpha = 2$ be the threshold of growth rate support.
Pattern $abc$ is discriminative since $GR(abc,D) = 2.4$.

\begin{definition}
\textbf{(Discriminative pattern).} Let $\alpha$ be a discriminative
threshold, $scr(p,D)$ be the discriminative score of pattern $p$ in
$D$. The pattern $p$ is discriminative if $scr(p,D) \geq \alpha$.
\end{definition}


In addition to the discriminative score, to evaluate the statistical significance
of a discriminative pattern we need to consider the confidence intervals ($CI$).
Confidence intervals are the result of a statistical
measure. They provide information about a range of values (lower
confidence interval ($LCI$) to upper confidence interval ($UCI$)) in
which the true value lies with a certain degree of probability.
$CI$ is able to assess the statistical significance of a result \cite{Morris1988}.
A confidence level of $95\%$ is usually selected. It means that
the $CI$ covers the true value in 95 out of 100 studies.

Let $a = |D_1(p)|$, $b = |D_1|- |D_1(p)|$, $c = |D_2(p)|$, $d = |D_2|-|D_2(p)|$, 
the $95\%$ $LCI$ and $UCI$ of $GR$ are estimated as formulas \ref{formular:lci-rg} and \ref{formular:uci-rg} repectively.

\begin{equation}
LCI_{GR} = e^{ \left(ln(GR) - 1.96\sqrt{\frac{1}{a} - \frac{1}{a+b} + \frac{1}{c} - \frac{1}{c+d} } \right)}
\label{formular:lci-rg}
\end{equation}

\begin{equation}
UCI_{GR} = e^{\left( ln(GR) + 1.96\sqrt{\frac{1}{a} - \frac{1}{a+b} + \frac{1}{c} - \frac{1}{c+d} } \right)}
\label{formular:uci-rg}
\end{equation}

Similarly, the $95\%$ $LCI$ and $UCI$ of $OR$ are estimated as formulas \ref{formular:lci-or} and \ref{formular:uci-or} repectively.

\begin{equation}
LCI_{ORS} = e^{ \left( ln(ORS) - 1.96\sqrt{\frac{1}{a} + \frac{1}{b} + \frac{1}{c} + \frac{1}{d} } \right)}
\label{formular:lci-or}
\end{equation}

\begin{equation}
UCI_{ORS} = e^{ \left( ln(ORS) + 1.96\sqrt{\frac{1}{a} + \frac{1}{b} + \frac{1}{c} + \frac{1}{d} } \right)}
\label{formular:uci-or}
\end{equation}

For example, consider the pattern $abc$ in the previous example,
the $95\% CI$ of $GR$ are $LCI_{GR}=0.37, UCI_{GR}={16.60}$.
Thus the $GR$ score of $abc$ is statistically significant because this 
score lies between $LCI$ and $UCI$ values.


\begin{definition}
\textbf{(Statistically significant discriminative pattern).}  Given a discriminance score $scr \in \{GR, ORS\}$, a 
discriminative threshold $\alpha$ and a lower confidence interval
threshold $\beta$, the pattern $p$ is statistically significant
discriminative in $D$ if $scr(p,D) \geq \alpha$ and $lci_{scr}(p, D)  > \beta$. 
\end{definition}

\textbf{Problem statement:} Given a two-class dataset $D$, a discriminance score $scr$ and two thresholds $\alpha$ and $\beta$, 
the problem is to discover the complete set of patterns $P$ that are statically significant discriminative for dataset $D$, discriminative measure $scr$, discriminative threshold $\alpha$ and lower confidence interval threshold $\beta$.

Note that this problem can be extended to discover all patterns which satisfy
multiple discriminative score thresholds and confidence intervals. In
particular, given a set of discriminative thresholds $\{SD = \alpha_1,
GR=\alpha_2, ORS=\alpha_3 \}$, and a set of lower confidence interval
thresholds  $\{LCI_{GR}=  \beta_1,  LCI_{ORS}=\beta_2\}$. We want to
discover all patterns which satisfy $SD \geq \alpha_1$ and $GR \geq
\alpha_2$ and $ORS  \geq  \alpha_3$ and $LCI_{GR}  >  \beta_1$ and
$LCI_{ORS} > \beta_2$.

In the example of Table \ref{Table:exData}, let $\alpha_1 = 0.2 ,  \alpha_2 = 2, \alpha_3 = 2$ be the 
thresholds of $SD$, $GR$, $ORS$ and $\beta_1=2$, $\beta_2=2$ be the lower confidence interval
thresholds, $abc$ is a
statistically significant discriminative pattern since
its scores satisfy these thresholds.


\section{Enumeration Strategy}
\label{sec:enum}
The main practical contribution of this paper is \sd, an efficient
algorithm for mining statistically significant discriminative patterns. 
This algorithm will be presented in the next section (Section \ref{sec:algo}).
\sd\ owes its efficiency to an original enumeration strategy of the patterns,
which allows to exploit some degree of anti-monotonicity on the measures
of discriminance and statistical significance.

The majority of enumeration strategies used in pattern mining algorithms
make a tree-shaped enumeration (called an {\em enumeration tree}) over
all the possible itemsets. This enumeration tree is based on
{\em itemset augmentation}: each itemset $p$ is represented by a node,
and the itemsets $p \cup \{e\}$ (for $e$ in $I$) are children of $p$:
the augmentation is the transition from $p$ to $p \cup \{e\}$. 
If such augmentation was conduced for all $e \in I$, this would lead
to enumerating multiple times the same itemset (ex: $ab \cup c = bc \cup a = abc$).
Each enumeration strategy imposes constraints on the $e$ that can be
used for augmentation at each step, preventing redundant enumeration
while preserving completeness. 
The other important component of pattern mining enumeration strategies
is the use of {\em anti-monotonicity properties}. 
When enumerating frequent itemsets, one can notice that if an
itemset $p$ is unfrequent ($sup(p, D) < min\_sup$), then no
super-itemsets $p' \supset p$ can be frequent
(necessarily $sup(p', D) < sup(p, D) < min\_sup$).
This allows to stop any further enumeration when an unfrequent
itemset $p$ is found, allowing a massive reduction in the
search space \cite{Agrawal1993}. As far as we know, no such
anti-monotonicity could be defined on measures of discriminance
or statistical significance. 

\begin{figure*}
\centering
\begin{forest}, baseline, qtree,
  [$\{\}$
    [\begin{tabular}{|c|} \hline 1 : -  \\ \hline  $abcfij$ \\  \hline \end{tabular}]
    [ \begin{tabular}{|c|} \hline 2 : - \\  \hline $abcegi$  \\ \hline \end{tabular}
		[\begin{tabular}{|c|} \hline 12 : - \\ \hline $abci$  \\ \hline \end{tabular}
			[\begin{tabular}{|c|} \hline 12 : 6 \\ \hline $bc$ \\  \hline \end{tabular}]
			[\begin{tabular}{|c|} \hline 12 : 7  \\ \hline $abc$ \\  \hline \end{tabular}]
			[\begin{tabular}{|c|} \hline 12 : 8  \\ \hline $bci$ \\  \hline \end{tabular}
				[\begin{tabular}{|c|} \hline 12 : 68 \\ \hline $bc$ \\  \hline \end{tabular}]			
				[\begin{tabular}{|c|} \hline 12 : 78  \\ \hline $bc$ \\  \hline \end{tabular}
	 				[\begin{tabular}{|c|} \hline 12 : 678 \\ \hline $bc$ \\  \hline \end{tabular} , edge={myedge}]
	 			]							
			]
			[\begin{tabular}{|c|} \hline 12 : 9  \\ \hline $a$ \\ \hline \end{tabular}]		
		]
		[\begin{tabular}{|c|} \hline 2 : 6  \\ \hline $bceg$ \\  \hline \end{tabular}]
		[\begin{tabular}{|c|} \hline 2 : 7  \\ \hline $abcg$ \\  \hline \end{tabular}
			[\begin{tabular}{|c|} \hline 2 : 67  \\ \hline $bcg$  \\  \hline \end{tabular},edge={myedge}]]
		[\begin{tabular}{|c|} \hline 2 : 8  \\ \hline $bcei$ \\  \hline \end{tabular}
			[\begin{tabular}{|c|} \hline 2 : 68  \\ \hline $dc$  \\  \hline \end{tabular},edge={myedge}]
			[\begin{tabular}{|c|} \hline 2 : 78  \\ \hline $bc$  \\  \hline \end{tabular},edge={myedge}
				[\begin{tabular}{|c|} \hline 2 : 678  \\ \hline $b$  \\  \hline \end{tabular}]
			]		
		]
		[\begin{tabular}{|c|} \hline 2 : 9  \\ \hline $aeg$  \\  \hline \end{tabular}
			[\begin{tabular}{|c|} \hline 2 : 69  \\ \hline $e$  \\  \hline \end{tabular},edge={myedge}]
			[\begin{tabular}{|c|} \hline 2 : 79  \\ \hline $a$  \\  \hline \end{tabular},edge={myedge}]
			[\begin{tabular}{|c|} \hline 2 : 89  \\ \hline $e$  \\  \hline \end{tabular},edge={myedge}]
		]
    ]
    [\begin{tabular}{|c|} \hline 3 : -  \\  \hline $abcfhj$ \\  \hline \end{tabular}
        [\begin{tabular}{|c|} \hline 13 : -  \\  \hline $abcfj$ \\  \hline \end{tabular}]
        [\begin{tabular}{|c|} \hline 23 : -  \\  \hline $abc$ \\  \hline \end{tabular}
            [\begin{tabular}{|c|} \hline 123 : -  \\  \hline $abc$ \\  \hline \end{tabular}
                [\begin{tabular}{|c|} \hline 123 : 7  \\  \hline $abc$ \\  \hline \end{tabular}
                    [\begin{tabular}{|c|} \hline 123 : 67  \\  \hline $bc$ \\  \hline \end{tabular} ]
                ]
            ]
        ]
    ]
  ]
\end{forest}
\caption{Tidset-itemset search tree}
\label{Fig:enumTree}
\end{figure*}
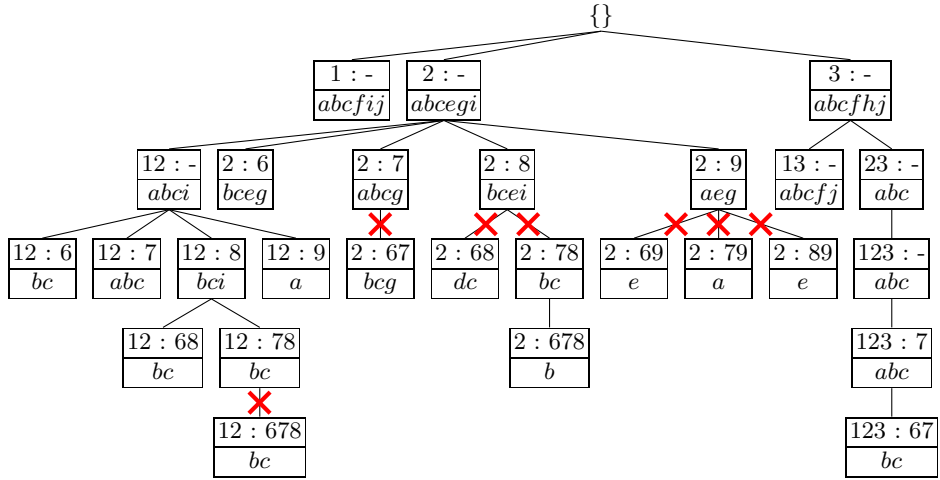

The enumeration strategy proposed in \sd\ also builds an enumeration tree. 
However, it is based on the tidsets and not the itemsets. 
Each node of the enumeration tree is a tidset (with the empty tidset at the root),
and the augmentation operation consists in adding a single tid:
the children of node of tidset $t$ are nodes of tidset $t \cup i$
for some $i \in \{1..n\}$. An example enumeration tree for the data
of Table \ref{Table:exData} is presented in Figure \ref{Fig:enumTree},
with the tidset written on the top of each node. Note that the tidset
is displayed with a separation of the tids from $D_1$ (case) and from $D_2$ (control).
For example, consider the node represented by \fbox{12 : 8}: this node
corresponds to the tidset $128$ in which $12$ are the positive tids,
and $8$ is the negative tid. The children of \fbox{12:8} are \fbox{12:68}
(augmentation by 6) and \fbox{12:78} (augmentation by 7).

Before delving deeper on the enumeration strategy that was used
to construct this tree, we explain how it is possible to recover
the itemsets (which are our expected outputs) from the tidsets.
This is a well known problem: itemsets and tidsets are in facts
dual notions, and they can be linked by two functions that form
a {\em Galois connection} \cite{Pasquier1999}. The main difference
in our definition is that the main dataset can be divided into two
parts ($D = D_1 \cup D_2$), and we want to be able to apply
functions of the Galois connection either in the complete dataset
$D$ or in any of its parts $D_1$ or $D_2$.

\begin{definition}[Galois connection]
For a dataset $D = D_1 \cup D_2$:
\begin{itemize}
	\item For any tidset $q \subseteq \{1..n\}$ and any itemset $p \subseteq I$, we define:
	
	\begin{equation*}
	    f(q, D) = \{ i \in I \; | \; \forall k \in q \;\; i \in t_k \}
	\end{equation*}
	\begin{equation*}
	    g(p, D) = \{ k \in \{1..n\} \;|\; p \subseteq t_k \}
	\end{equation*}

	\item For any tidset $q_1 \subseteq D_1.tids$ and any itemset $p \subseteq I$, we define: 
	\begin{equation*}
	    f_1(q_1, D_1) = \{ i \in I \; | \; \forall k \in q_1 \;\; i \in t_k \}
	\end{equation*}
	\begin{equation*}
	    g_1(p, D_1) = \{ k \in D_1 \;|\; p \subseteq t_k \}
	\end{equation*}

	\item  For any tidset $q_2 \subseteq D_2.tids$ and any itemset $p \subseteq I$, we define:
	\begin{equation*}
	    f_2(q_2, D_2) = \{ i \in I \; | \; \forall k \in q_2 \;\; i \in t_k \}
	\end{equation*}
	\begin{equation*}
	    g_2(p, D_2) = \{ k \in D_2 \;|\; p \subseteq t_k \}
	\end{equation*}

\end{itemize} 
\end{definition}

Note that this definition marginally differs from the standard
definition presented in \cite{Pasquier1999}: here for convenience
we operate on the set of tids $\{1..n\}$, whereas the standard
definition operates on the set of transaction $\{t_1,...,t_n\}$. 

In Figure \ref{Fig:enumTree}, under each tidset $q$, its associated
itemset $f(q, D)$ is displayed. For example for node $\fbox{12:8}$,
the itemset $f(128, D) = bci$ is displayed. One can verify in
Table \ref{Table:exData} that $bci$ is the only itemset common
to the transactions $t_1$, $t_2$ and $t_8$.

Finding an itemset associated to a tidset is a trivial use of
the Galois connection. A more advanced use is to define a
{\em closure operator}, which takes as input any tidset $q$,
and returns the closed pattern that has the smallest tidset containing $q$. 

\begin{definition}[Closure operator]
For a dataset $D$ and any tidset $q \subseteq \{1..n\}$, the closure operator is defined as:
$$ c(q, D) = g \circ f(q, D) $$
The output of $c(q, D)$ is the tidset of the closed itemset having the smallest tidset containing $q$.
\end{definition}

We can similarly define $c_1(q_1, D_1) = g_1 \circ f_1(q_1, D_1)$ for $q_1 \subseteq D_1.tids$ and $c_2(q_2, D_2) = g_2 \circ f_2(q_2, D_2)$ for $q_2 \subseteq D_2.tids$.

Note that the standard literature on pattern mining defines the
closure operator as taking an itemset as input, whereas here we
define it as having a tidset as input. Replacing $g \circ f$ by
$f \circ g$ gives the dual closure operator taking itemsets as input.

The basics of the enumeration have been given: the enumeration
proceeds by augmenting tidsets (starting from the empty tidset),
and for each tidset function $f$ of the Galois connection gives
the associated itemset. The specificity of our enumeration strategy
is to be designed around statistically significant discriminative patterns. 
This appears first in our computation of closure: we divide the
computation of closure in the two sub-datasets $D_1$ and $D_2$. 
This intermediary step allows some early pruning.
Second, most measures of discriminance require the pattern to have
a non-zero support in $D_2$ ($GR$ and $ORS$). The same condition
apply for measures of statistical significance: in both cases we
need to defer measures of interest of patterns until it has some tids in $D_2$.
Our enumeration strategy thus operates in two steps:
\begin{enumerate}
	\item From the empty set, it enumerates closed tidsets containing
	only elements of $D_1$ (case group). 
	\item For each of those tidset containing only tids of $D_1$,
	augmentations using only tids of $D_2$ are generated and their
	closure is computed. Any subsequent augmentation of such nodes
	will only be allowed to be augmented by tids of $D_2$.
\end{enumerate}

More formally, let $q \subseteq \{1..n\}$ be a tidset,
with $q = q^+ \cup q^-$, where $q^+ \subseteq D_1.tids$ and
$q^- \subseteq D_2.tids$. Then the possible augmentations of $q$ are:
\begin{itemize}
	\item ({\em Rule 1}) if $q^- = \emptyset$: $q$ can either:
	\begin{itemize}
		\item ({\em Rule 1a}) be augmented with $k \in D_1.tids$ such that $k < min(q^+)$
		\item ({\em Rule 1b}) be augmented with $k \in D_2.tids$
	\end{itemize} 
	\item ({\em Rule 2}) if $q^- \neq \emptyset$: $q$ can only be augmented with tid $k \in D_2.tids$ such that $k < min(q^-)$
\end{itemize}

This enumeration mechanic is based on imposing an arbitrary ordering
on the tidsets, a classical technique when enumerating itemsets. 
It is guaranteed to avoid enumerating duplicates. 

More interestingly, we show that it allows to benefit from an anti-monotony
property on the measures of statistical significance and discriminance.

\begin{theorem}[Anti-monotonicity]
	\label{th:antimonotonicity}
	Let $q_1$ and $q_2$ be two tidsets such as: $q_1^+ = q_2^+$ and $q_1^- \subset q_2^-$ (we have $q_1^+ \neq \emptyset$ and $q_2^- \neq \emptyset$). Let $p_1 = f(q_1, D)$ and $p_2 = f(q_2, D)$.
	Then: 
	\begin{enumerate}
		\item $scr(p_1, D) > scr(p_2, D)$ with $scr$ a discriminance measure in $\{SD, GR, ORS\}$.
		\item $lci(p_1, D) > lci(p_2, D)$ with $lci$ a lower confidence interval in $\{LCI_{ORS}, LCI_{GR}\}$.
	\end{enumerate}
\end{theorem}

\textbf{Proof}:
1) For the tidset $q_1$, let $a = |q_1^+|$ be the number of positive tids and $c = |q_1^-|$ be the number of negative tids ($0 \leq a \leq|D_1|$, $0 \leq c \leq |D_2|$). Let $b = |D_1|-a$, and $d = |D_2|-c$. Then $SD$, $GR$, and $ORS$ of $p_1$ are estimated as follows:

\begin{equation*}
SD(p_1,D)  = \frac{a}{a+b}-\frac{c}{c+d}
\end{equation*}

\begin{equation*}
GR(p_1,D) = \frac{a/(a+b)}{c/(c+d)}
\end{equation*}

\begin{equation*}
ORS(p_1,D)  =  \frac{a.d}{b.c}
\end{equation*}

We have $q_1 \subset q_2$, then $|q_1| - |q_2| = x > 0 $, where by definition of $q_1$ and $q_2$ those $x$ tids are part of $D_2.tids$.
$SD$, $GR$, and $ORS$ of $p_2$ are thus estimated as follows:

\begin{equation*}
SD(p_2,D)  = \frac{a}{a+b}-\frac{c+x}{c+d} < SD(p_1,D)
\end{equation*}

\begin{equation*}
GR(p_2,D) = \frac{a/(a+b)}{(c+x)/(c+d)} < GR(p_1,D)
\end{equation*}

\begin{equation*}
ORS(p_2,D)  = \frac{a.(d-x)}{b.(c+x)} < ORS(p_1,D)
\end{equation*}

2) Please refer to the supporting document for the detailed demonstration of this part. 

\hfill $\square$

This theorem provides pruning by anti-monotonicity in our
enumeration strategy: for a node having a tidset with tids
both from $D_1.tids$ and $D_2.tids$, if the discriminance or statistical
significance measures are below a given threshold, then necessarily
its augmentations will also be under the threshold. 
Hence this part of the enumeration tree can be pruned. 

Consider the node \fbox{2:8} for example. Its associated itemset
is $bcei$ and $ORS(bcei,D) = 3/4$. If the threshold is 2, then this
node can be pruned and its augmentations need not be computed.
This allows to significantly reduce the search space.

\section{SSDPS: Algorithm Design and Implementation}
\label{sec:algo}
In this section, we present the \sd\ algorithm. We present
in details how the enumeration strategy presented in Section \ref{sec:enum}
is exploited in the algorithm.
We then show several techniques to improve the performance of the algorithm.

\subsection{Algorithm Design}

As mentioned in the previous section, our algorithm is based on an
enumeration of the tidsets. It discovers statistically significant
discriminative closed patterns. 
The main procedure for enumerating tidsets is given in Algorithm
\ref{alg:main}. This procedure calls the recursive procedure
$positive\_expand$ (Algorithm \ref{alg:posexpand}) to find closed
frequent itemsets in the positive class. Computing discriminative
patterns relies on the recursive procedure $negative\_expand$
(Algorithm \ref{alg:negexpand}).

Delving more into details, $positive\_expand$ (Algorithm \ref{alg:posexpand})
is based on the principles of the LCM algorithm \cite{Uno2004}, the
state of the art for mining closed frequent itemsets.
$positive\_expand$ takes as input the tidset $t$ of a pattern that
is closed in $D_1$ and a tid $e \in D_1.tids$ that can be used to augment $t$.
This augmentation is performed on line \ref{alg:posexp:augm}, and
the pattern $p$ associated to the augmented tidset $t^+ = t \cup \{e\}$
is computed in line \ref{alg:posexp:getpatt}.
If $p = \emptyset$, there are no items common to all transactions
of $t^+$ so the enumeration can stop (test of line \ref{alg:posexp:testPnoEmpty}).
Else, we can continue the enumeration by applying {\em Rule 1} of
enumeration presented in Section \ref{sec:enum}.
Lines \ref{alg:posexp:compClo} to \ref{alg:posexp:1stParentOK} apply
the LCM principles of enumerating closed itemsets without redundancies
(the interested reader in referred to \cite{DBLP:journals/is/LeroyKTA17}
Section 3.2 for a recent description of these principles).
At this step of the enumeration, the closure is computed in $D_1$ (line \ref{alg:posexp:compClo}). 
The test of line \ref{alg:posexp:cloExtends} verifies if the closure
actually extends the tidset, requiring a further verification in line
\ref{alg:posexp:1stParentOK}, and the construction of the new extended
tidset (line \ref{alg:posexp:newtidset}).

Lines \ref{alg:posexp:loopAugmPos} to \ref{alg:posexp:recCallPos1} implement
{\em Rule 1a} of enumeration, allowing to grow the positive part of the tidset.
Lines \ref{alg:posexp:loopAugmNeg} to \ref{alg:posexp:recCallNeg1} implement
{\em Rule 1b} of enumeration, stopping the growth of the positive part and
starting to grow the negative part of the tidset. 
The same logic is followed in lines \ref{alg:posexp:redData2} to \ref{alg:posexp:negExp2},
in the case where the tidset is not extended by the closure (test of line \ref{alg:posexp:1stParentOK} is false).

The final expansion of the tidset is handled by $negative\_expand$ (Algorithm \ref{alg:negexpand}),
that can only perform augmentations with negative tidsets. 
It is very similar to $positive\_expand$, with several key differences.
The first obvious one is that the closure is this time computed in $D_2$ (line \ref{alg:negexp:computeClo}).
The second one is that only {\em Rule 2} of enumeration can be applied (lines \ref{alg:negexp:recCall1}
and \ref{alg:negexp:recCall2}). The third and most important difference is that because
we have tidsets with positive and negative tids, we can compute discriminance as well
as statistical significance measures. Hence, Theorem \ref{th:antimonotonicity} can be
applied to benefit from pruning by anti-monotonicity. This is done in line \ref{alg:negexp:pruning}.

As an example of the execution of the algorithm, consider tidset $12$.
Its associated itemset is $abci$ and its closure in $D_1$ is $12$.
Thus $abci$ is closed in $D_1$. Then $12$ will be combined with all
tids in $D_2.tids$ to find discriminative patterns. In particular, the following
tidsets are created: $126$, $127$, $128$, and $129$.
Consider the tidset of $128$. We have $f(128,D) = bci$ and $c_2(128,D_2) = 128$.
Thus $bci$ is closed in $D_2$. The discriminative scores of $bci$ in $D$
are: $ORS(bci,D)  =  2$,  $GR(bci,D) =  1.6$, $SD(bci,D) = 0.15$.
Suppose the discriminative thresholds are: $ORS = 1.5$, $GR = 1.5$ and
$SD = 0.1$. $bci$ is a discriminative pattern since it satisfies all
given thresholds, and $128$ is the tidset containing $bci$.
In contrast, $1278$ does not satisfy discriminative thresholds. Thus
all branches expanded from this node are pruned.

The \sd\ algorithm can discover patterns even from small tidset
(upper nodes of the enumeration tree). It means that the patterns
with very low support are taken into consideration by the \sd\ algorithm.

\begin{algorithm}
\caption{Exhaustive search algorithm}
Input: two-class dataset $D$, discriminative thresholds $\alpha$, confidence intervals $\beta$\\
Output: the set of statistically significant discriminative patterns\\

\begin{algorithmic}[1]
\State transaction id set $t= \emptyset $
\For {each transaction id $e$ in $positive\_class$} \label{alg:mainloop}
\State $positive\_expand(t, e, D, \alpha, \beta)$
\EndFor
\end{algorithmic}
\label{alg:main}
\end{algorithm}

\begin{algorithm}[h]
\caption{Positive class expanding}
Procedure $positive\_expand(t,e,D,\alpha,\beta)$\\
\begin{algorithmic}[1]
\State $t^+ \gets t \cup \textit{\{e\}}$ \label{alg:posexp:augm}
\State $p \gets f(t^+,D) $    \label{alg:posexp:getpatt}  
\If {$p$ is not empty} \label{alg:posexp:testPnoEmpty}
\State $t\_ext^+ \gets c_1(t^+,D_1)$  \label{alg:posexp:compClo}  
\If {$t\_ext^+ \neq t^+$ } \label{alg:posexp:cloExtends}
	\If {$max(t\_ext^+) < e$} \label{alg:posexp:test1stParent}
		\State $q \gets t^+ \cup t\_ext^+$ \label{alg:posexp:newtidset}
		\State $RD \gets reduced\_dataset(q,D)$ \label{alg:posexp:datasetred}
		\For {each $e^+$ in $D_1.tids \setminus q$} \label{alg:posexp:loopAugmPos}
			\If {$e^+ < e$} \label{alg:posexp:1stParentOK} \State $positive\_expand (q,e^+,RD,\alpha, \beta)$ \label{alg:posexp:recCallPos1} \EndIf
		\EndFor
		\For {each $e^-$ in $D_2.tids$} \label{alg:posexp:loopAugmNeg} \State $negative\_expand (q,e^-,RD,\alpha, \beta)$ \label{alg:posexp:recCallNeg1} \EndFor
	\EndIf
\Else
	\State $RD \gets reduced\_dataset(t^+,D)$ \label{alg:posexp:redData2}
	\For {each $e^+$ in $D_1.tids$}
		\If {$e^+ < min(t^+)$} \State $positive\_expand (t^+,e^+,RD, \alpha, \beta)$ \EndIf
	\EndFor
	\For {each $e^-$ in $D_2.tids$} \State $negative\_expand (t^+,e^-,RD,\alpha, \beta)$ \label{alg:posexp:negExp2} \EndFor
\EndIf
\EndIf
\end{algorithmic}
\label{alg:posexpand}
\end{algorithm}

\begin{algorithm}[h]
\caption{Negative class expanding}
Procedure $negative\_expand(t,e,D,\alpha,\beta)$\\
\begin{algorithmic}[1]
\State $t^- \gets t \cup \{e\}$
\State $p \gets f(t^-,D) $ 
\If {$p \neq \emptyset$}
	\If{$check\_significance(p,D,\alpha,\beta)$ is true} \label{alg:negexp:pruning}
		\State $t\_ext^- \gets c_2(t^-,D_2)$ \label{alg:negexp:computeClo} 
			\If {$t\_ext^- \neq t^-$}
				\If {$max(t\_ext^-) < e$}
					\State $q \gets t^- \cup t\_ext^-$
					\State $q\_ext \gets c(q,D) $
					\State $p' \gets f(q,D) $
					\If{$q\_ext$ = $q$}
						\If{$check\_significance(p',D,\alpha,\beta)$ is true}
							\State output: $p'$
						\EndIf
					\EndIf						
					\State $RD \gets reduced\_dataset(q,D)$
					\For {each $e^- \in D_2.tids \setminus q$} 
						\If {$ e^- < e $}\State $negative\_expand (q,e^-,RD,\alpha, \beta)$ \label{alg:negexp:recCall1} \EndIf \EndFor
				\EndIf
			\Else
			\State $t\_ext \gets c(t^-,D)$
			\If{$t\_ext=t^-$} 
				\State output: $p$ 
			\EndIf				
			\State $RD \gets reduced\_dataset(t^-,D)$
			\For {each $e^- \in D_2.tids \setminus t^- $}
				\If {$e^- < e$} \State $negative\_expand (t^-,e^-,RD,\alpha, \beta)$ \label{alg:negexp:recCall2} \EndIf \EndFor
			\EndIf
	\EndIf
\EndIf
\end{algorithmic}
\label{alg:negexpand}
\end{algorithm}

\subsection{Implementation}
The performance of the \sd\ algorithm relies on the computation
of 2 functions: $f()$ (compute associated itemset of a tidset)
and $c()$ (compute closure operator of a tidset). Both functions
need to compute the intersection of two sets. With integer data
presentation this operator spends $O(max(n,m))$ iterations,
where $n$ and $m$ are the size of the two sets. Thus, the time
required for each task of computing associated itemset (or closure
operator) is $O(I*max(n,m))$, where $I$ is the number of items in dataset.
In this study, we use the dataset reduction technique \cite{Han2000}
to decrease the number of rows, i.e. the number of items $I$
(function $reduced\_dataset$). With the use of this technique,
the number of items is significantly reduced after each step of searching.

In addition, the \sd\ algorithm uses \textit{vertical} data format \cite{Pan2003,zaki2002}
combined with a binary data representation to improved its performances.
In this format, each row represents an item and columns correspond to tids.
The value 1 at position $(i,j)$ indicates that the item $i$ is presents in
the transaction having tid $j$. In contrast, 0 indicates that item $i$ is
absent in the transaction having tid $j$.
The benefits of this data representation are:
\begin{itemize}
    \item The task of computing support is simpler and faster.
We only need tidset to compute the support of an itemset.
    \item The vector of bits (bitset) representation allows to efficiently
compute support of itemsets by using bit operations such as AVX2.
    \item We can easily distinguish the positive and negative
tids in a tidset. This helps us to estimate the discriminative
scores and confidence intervals effectively.
\end{itemize}

\section{Experimental Results}
\label{sec:expes}
This section presents various experiments
to evaluate the performance of the \sd\ algorithm.
In addition, we apply the \sd\ to discover multiple SNPs combinations in a real genomic dataset. 
All experiments have been conducted on a laptop with
Core i7-4600U CPU @ 2.10GHz, 16GB memory and Linux operating system.

\subsection{Synthetic data}
A synthetic two-class data was created to evaluate the pruning
strategy as well as compare \sd\ with other algorithms.
This dataset includes 100 transactions (50 transactions for each class).
Each transaction contains 262 items which are randomly set by value 0 or 1.
The density of data is set up to $33\%$.

\subsection{Pruning Efficiency Evaluation}
To evaluate the pruning efficiency of the \sd\ algorithm,
we executed 2 setups on the synthetic dataset.
\begin{itemize}
    \item Setup 1: use $OR$ as discriminative measure; the discriminative threshold $\alpha = 2$.
    \item Setup 2: use $OR$ as discriminative measure and $LCI$ of $OR$ as statistically significant testing; the discriminative threshold $\alpha = 2$, and LCI threshold $\beta=2$.
\end{itemize}{}

As the result, the running time and the number of output
patterns significantly reduce when applying $LCI_{ORS}$.
In particular, with the setup 1, the \sd\ algorithm
generates 179,334 patterns in 38.69 seconds
while the setup 2 returns 18,273 patterns
in 9.10 seconds. This result shows that
a large amount of patterns is removed by
using statistically significant testing.

\subsection{Comparison with Existing Algorithms}
We compare the performance of the \sd\ algorithm with
two well-known algorithms: CIMCP \cite{Guns2011} and SFP-Growth \cite{Ma2010}.
Note that these algorithms deploy
discriminative measures which are different from the measures of \sd\ .
In particular, CIMCP uses one of measures such as chi-square, information-gain and
gini-index as a constraint to evaluate discriminative patterns while SFP-GROWTH applies $-log(p\_value)$.
For this reason, the number of output patterns and the running
times of these algorithms should be different.
It is hence not fair to directly compare the performance of \sd\ with these algorithms. However, to have an initial comparison of the performance as well as the quantity of discovered patterns, we select these algorithms.


We ran three algorithms on the same synthetic data.
The used parameters and results are given in Table \ref{Table:alParameters}.

\begin{table}[]
\centering
\caption{Used parameters and results of 3 algorithms}
\label{Table:alParameters}
\begin{tabular}{|l|l|l|r|r|}
\hline
\textbf{Algorithms} & \textbf{Measure} & \textbf{Threshold} & \textbf{\#Patterns} & \textbf{Time(seconds)} \\ \hline
\sd &  $OR$, $LCI\_ORS$
    & $\alpha = 2$, $\beta = 2$
    & 49,807
    & 73.69\\ \hline
CIMCP   & Chi-square 
        & 2
        & 5,403,688
        & 143\\ \hline
SFP-GROWTH & 
        -log(p$\_$value)
        & 3
        & *
        &  $>172$ (out of memory) \\ \hline
\end{tabular}
\end{table}

As the result, the \sd\ algorithm finds 49,807 patterns in 73.69 seconds;
CIMCP discovers 5,403,688 patterns in 143 seconds.
The SFP-GROWTH runs out of storage memory after 172 seconds.
Hence the number of patterns isn't reported in this case.

In comparison with these algorithms the \sd\ gives a comparable performance, while the number of output patterns is much smaller.
The reason is that the output patterns of \sd\ are tested for statistical significance by $CI$ while other algorithms use only the discriminative measure.
However, this amount of patterns is also larger for real biological analysis. Thus, searching for a further reduced number of significant patterns should be taken into account.



\subsection{Experiment on Real Genetic Data}
In this experiment, we apply \sd\ to find
multiple SNPs combinations from Age-Related Macular Degeneration (AMD) dataset \cite{Klein2005}. This dataset includes
103,611 SNPs of 96 case individuals and 50 control individuals.
Based on GWAS analysis which is given in \cite{Klein2005}, 
there are 2 SNPs (rs1329428, rs380390) that are
reported as association with AMD.
The aim of this experiment is to study whether or not
the \sd\ algorithm can discover these SNPs.

In genetic variant datasets, each SNP has three genotypes
which are here considered as the items.
The case group and the control group are equivalent to the positive class and the negative class, respectively.
Since the amount of genotypes is very
large, using all genotypes to find combinations is infeasible. In
addition, many individual genotypes may not really meaningful. For example, the genotypes have very high frequency or that occur
more in the control group than in the case group.
These genotypes are considered as noise since they can be combined
with any patterns without decreasing discriminative score.
Thus, discarding these genotypes is necessary.

To effectively search multiple SNPs combinations, we
propose to use a heuristic strategy as follow:
\begin{itemize}
    \item Step 1: using $p\_value$ and support of genotype in the control group (denoted $control\_support$) to select candidate genotypes. In particular, if $a$ is the $p\_value$ threshold and $b$ the control support threshold, we select genotypes which have $p\_value \leq a$ and $control\_support \leq b$. The reason is that, the $p\_value$ guarantees that the selected candidates are significant, while the control support is used to eliminate very common genotypes.
\item Step 2: discovering SNPs combinations by using the \sd\ algorithm.
\end{itemize}

To evaluate the proposed approach, we use 2 sets
of parameters as follows:
\begin{itemize}
\item Set 1: a fixed $p\_value$ threshold at 0.001 and three $control\_support$ thresholds: $30\%, 50\%$ and $70\%$.
\item Set 2: a fixed $control\_support$ at $30\%$ and three $p\_value$ thresholds: $0.005$, $0.01$, and $0.05$.
\end{itemize}

As the results, patterns including SNP rs1329428 and rs380390
are reported in all cases. Table \ref{Table:reAMD_set1}
summarizes the results of the \sd\ algorithm with parameters
tuned according to Set 1 (variation of $control\_support$).

\begin{table}
  \centering
  \caption{Pattern generated on AMD dataset with different control support}
  \label{Table:reAMD_set1}
  \begin{threeparttable}
    \begin{tabular}{llllll}
      \hline
      Support & rs1329428\_2    & rs380390\_0     & Both    & Number of patterns   & Running Time(s) \\
      \hline
      30\%    & 21   & 5   & 5   & 29   &  16  \\
      50\%    & 59   & 9   & 9   & 299  & 145  \\
      70\%    & 45   & 2   & 2   & 307  & 287  \\
      \hline

    \end{tabular}

    \footnotesize
    \begin{tablenotes}[para,flushleft]
      
      rs1329428$\_$2, rs380390$\_0$: the number of patterns containing
      these SNPs. Both: the number of patterns including both SNPs.
      
    \end{tablenotes}
  \end{threeparttable}
\end{table}

Table \ref{Table:reAMDSet2} summarizes the results of the \sd\
algorithm with parameters tuned according to Set 2 (variation of  the $p\_value$).
Again, in all cases, patterns including the two interesting SNPs are
output. Furthermore, the total number of output patterns is limited,
whatever the $p\_value$. However, the execution times are more important.
This is mainly due to the number of selected SNPs during the filtering step.

\begin{table}
  \centering
  \caption{Patterns generated on AMD dataset for different p\_values}
  \label{Table:reAMDSet2}
  \begin{threeparttable}
    \begin{tabular}{llllll}
      \hline
      P\_value & rs1329428\_2 & rs380390\_0 & Both & Number of Patterns & \begin{tabular}[c]{@{}l@{}}Running Time(s)\end{tabular} \\
      \hline
      0.005    & 22        & 4        & 4    & 35      &  120                                                      \\
      0.01     & 25        & 5        & 5    & 46      &  465                                                      \\
      0.05     & 25        & 3        & 3    & 51      & 1750                                              
      \\
      \hline         

    \end{tabular}

    \footnotesize
    \begin{tablenotes}[para,flushleft]
      rs1329428$\_$2, rs380390$\_0$: the number of patterns containing
      these SNPs. Both: the number of patterns including both SNPs.
    \end{tablenotes}
  \end{threeparttable}
\end{table}

\section{Conclusion and Perspectives}
\label{sec:conc}
In this paper we propose a novel algorithm, called \sd, that efficiently discover
statistically significant discriminative patterns from a two-class dataset.
The algorithm directly uses discriminative measures and confidence intervals as
anti-monotonic properties to efficiently prune the search space.
Experimental results show that the performance
of the \sd\ algorithm is better than other discriminative pattern mining algorithms.
However, the number of patterns generated by  \sd\
is still large for manual analysis.
To discover high-order SNP combinations
on real genomic data, a heuristic approach was proposed.
Interesting patterns were discovered by this approach. 
Many of them include SNPs 
which are known as association with the given
disease.
However, choosing appropriate thresholds to select individual genotypes
is still difficult, and requires a good expertise from the users.
Our first perspective is to investigate methods to suggest good
thresholds to the user based on characteristics of the dataset.
Another perspective is to apply multiple hypothesis testing in order to further remove uninteresting patterns.


\bibliographystyle{IEEEtran}


\pagebreak
\section*{Support document}
In this support document, we present details of the demonstration of theorem 1.2. Let recall the presence and absence of pattern $p$ in $D$. It is presented by a 2x2 contingency table as follow:

\begin{table}[h]
\centering
\caption{A 2x2 contingency table of a pattern in case-control data}
\label{Table:1}
\begin{tabular}{| l | c | c | c | }
\hline
 & Presence & Absence & Total \\ 
 \hline
 Case & $a$ & $b$ & $|D_1|$ \\  
 \hline
 Control & $c$ & $d$ & $|D_2|$ \\
 \hline
\end{tabular}
\end{table}

Let $q_i - g(q_i,D)$ and $q_j - g(q_j,D)$ be two TI-pairs in the same equivalent class. We have $q_i \subset q_j$ and $p_i = g(q_i,D)$, $p_j = g(q_j,D)$. Let $|q_j| - |q_i| = 1$ be a minimal difference between $q_j$ and $q_i$ we have:

The lower confidence intervals of $ORS$ of $p_i$ ans $p_j$ are given:

\begin{equation}
LCI\_ORS(p_i,D) = \exp \left( ln(\frac{ad}{bc}) - 1.96 \sqrt{\frac{1}{a} + \frac{1}{b} + \frac{1}{c} + \frac{1}{d}} \right)
\end{equation}

\begin{equation}
LCI\_ORS(p_j,D) = \exp \left( ln(\frac{a(d-1)}{b(c+1)}) - 1.96 \sqrt{\frac{1}{a} + \frac{1}{b} + \frac{1}{c+1} + \frac{1}{d-1}} \right)
\end{equation}

The lower confidence intervals of $GR$ of $p_i$ and $p_j$ are given:

\begin{equation}
LCI\_GR(p_i,D) = \exp \left( ln(\frac{a(c+d)}{c(a+b)}) - 1.96 \sqrt{\frac{1}{a} - \frac{1}{a+b} + \frac{1}{c} - \frac{1}{c+d}} \right)
\end{equation}

\begin{equation}
LCI\_GR(p_j,D) = \exp \left( ln(\frac{a(c+d)}{(c+1)(a+b)}) - 1.96 \sqrt{\frac{1}{a} - \frac{1}{a+b} + \frac{1}{c+1} - \frac{1}{c+d}} \right)
\end{equation}

For all integers $a,b,c > 0$ and all integers $d>1$ we want to demonstrate that: $(1) > (2)$ and $(3) > (4)$.

\textbf{-Demonstrate $LCI\_ORS(p_i,D) > LCI\_ORS(p_j,D)$}

The lower confidence interval of $ORS$ of $p_i$ and $p_j$ are given:

\begin{equation*}
LCI\_ORS(p_i,D) = \exp \left( ln(\frac{ad}{bc}) - 1.96 \sqrt{\frac{1}{a} + \frac{1}{b} + \frac{1}{c} + \frac{1}{d}} \right)
\end{equation*}

\begin{equation*}
LCI\_ORS(p_j,D) = \exp \left( ln(\frac{a(d-1)}{b(c+1)}) - 1.96 \sqrt{\frac{1}{a} + \frac{1}{b} + \frac{1}{c+1} + \frac{1}{d-1}} \right)
\end{equation*}

First of all we can rewrite some terms and give their bounds.

\begin{equation*}
\alpha = \frac{1}{a} + \frac{1}{b} \quad \textrm{so} \quad 0 < \alpha \leq 2
\end{equation*}

\begin{equation*}
0 < \frac{1}{d} \leq \frac{1}{2}, \quad 0 < \frac{1}{d-1} \leq 1
\end{equation*}

\begin{equation*}
0 < \frac{1}{c} \leq 1, \quad 0 < \frac{1}{c+1} \leq \frac{1}{2}
\end{equation*}

Now, we calculate the difference:

\begin{equation*}
LCI\_ORS(p_i,D) - LCI\_ORS(p_j,D) = ln \frac{d(c+1)}{c(d-1)} + 1.96 \left( \sqrt{\alpha + \frac{1}{c+1} + \frac{1}{d-1}} - \sqrt{\alpha + \frac{1}{c} + \frac{1}{d} } \right)
\end{equation*}

The first term is clearly positive, the last one is the hardest to treat. With a little trick we can give another expression for this difference:\\

\begin{equation*}
\sqrt{\alpha + \frac{1}{c+1} + \frac{1}{d-1}} - \sqrt{\alpha + \frac{1}{c} + \frac{1}{d} } 
\end{equation*}

\begin{equation*}
= \frac{ \frac{1}{c+1} - \frac{1}{c} + \frac{1}{d-1} - \frac{1}{d} }{ \sqrt{\alpha + \frac{1}{c+1} + \frac{1}{d-1}} + \sqrt{\alpha + \frac{1}{c} + \frac{1}{d} } }
\end{equation*}

\begin{equation*}
= \frac{ \frac{-1}{c(c+1)} + \frac{1}{d(d-1)} } { \sqrt{\alpha + \frac{1}{c+1} + \frac{1}{d-1}} + \sqrt{\alpha + \frac{1}{c} + \frac{1}{d} } }
\end{equation*}

The denominator is always positive. We can notice that if $d \leq c+1$ then the numerator is also positive so $LCI\_ORS(p_i,D) > LCI\_ORS(p_j,D)$. We must treat the other case.

Let us suppose $d \geq c+2$. Let us rewrite the difference:

\begin{equation*}
LCI\_ORS(p_i,D) - LCI\_ORS(p_j,D) = \ln \frac{d(c+1)}{c(d-1)} -1.96 \frac{ \frac{1}{c(c+1)} - \frac{1}{d(d-1)} } { \sqrt{\alpha + \frac{1}{c+1} + \frac{1}{d-1}} + \sqrt{\alpha + \frac{1}{c} + \frac{1}{d} } }
\end{equation*}

In this case we know that the fraction is strictly positive, so we have to maximize it to lower bound the difference. We can remove some terms:

\begin{equation*}
LCI\_ORS(p_i,D) - LCI\_ORS(p_j,D) \geq \ln \frac{d(c+1)}{c(d-1)} - 2 \frac{ \frac{1}{c(c+1)} } { \sqrt{\frac{1}{c+1} + \frac{1}{d-1}} + \sqrt{\frac{1}{c} + \frac{1}{d} } }
\end{equation*}

It gives a general expression for the lower bound. The problem is it depends on two variables, so the idea is to removed $d$. We get quickly:

\begin{equation*}
\ln \frac{d(c+1)}{c(d-1)} = \ln \left( 1+ \frac{1}{d-1} \right) + \ln \left( 1 + \frac{1}{c} \right) \geq \ln \left( 1+ \frac{1}{c} \right)
\end{equation*}

Moreover

\begin{equation*}
\sqrt{\frac{1}{c+1} + \frac{1}{d-1}} + \sqrt{\frac{1}{c} + \frac{1}{d} } \geq \sqrt{\frac{1}{c+1}} + \sqrt{\frac{1}{c}}
\end{equation*}

Then we get,

\begin{equation}
LCI\_ORS(p_i,D) - LCI\_ORS(p_j,D) \geq \ln \left(1+\frac{1}{c}\right) - 2 \frac{ \frac{1}{c(c+1)}} {\sqrt{\frac{1}{c+1}}  + \sqrt{\frac{1}{c}} }
\end{equation}

This lower bound depends only on $c$ but studying directly this function is not simple. That's why, we can first simplify it.

\begin{equation*}
LCI\_ORS(p_i,D) - LCI\_ORS(p_j,D) \geq \ln \left(1+\frac{1}{c}\right) - 2 \frac{ \frac{1}{c(c+1)}}   {2 \sqrt{\frac{1}{c+1}}}
\end{equation*}

\begin{equation*}
\geq \ln \left(1+\frac{1}{c}\right) - \frac{1}{c \sqrt{c+1}}
\end{equation*}

\begin{equation*}
\geq \ln \left(1+\frac{1}{c}\right) - \frac{1}{c \sqrt{c}}
\end{equation*}

Let us introduce the function $f$ defined by:

\begin{equation*}
\forall x > 0, f(x) = \ln \left(1+\frac{1}{x}\right) - \frac{1}{x \sqrt{x}}
\end{equation*}

We can derive this function

\begin{equation*}
f'(x) = \frac{-\frac{1}{x^2}}{1+\frac{1}{x}} + \frac{3}{2} \frac{1}{x^2\sqrt{x}}
\end{equation*}

\begin{equation*}
= \frac{-1}{x(x+1)} + \frac{3}{2} \frac{1}{x^2\sqrt{x}}
\end{equation*}

\begin{equation*}
= \frac{-2x \sqrt{x} + 3(x+1)}{2x^2 \sqrt{x}(x+1)}
\end{equation*}

This denominator is always positive. Let us look at $ -2x \sqrt{x} + 3(x+1) \leq 0 $:

\begin{equation*}
-2x \sqrt{x} + 3(x+1) \leq 0 \longleftrightarrow 3 \leq x(2\sqrt{x}-3) 
\end{equation*}

The function $x \mapsto x(2\sqrt{x}-3) $ is clearly a growing function. As when $x=4$ the inequality is true, it is true for all $x \geq 4$. It shows that $f'$ is negative on $[4,+\infty]$. So $f$ is decreasing on the same interval. However $lim_{x \rightarrow +\infty}f(x)=0$. Hence, we know that $LCI\_ORS(p_i,D) \geq LCI\_ORS(p_j,D)$ for all $c \geq 4$.

The three cases $c=1,2$ and 3 have finally to be treated, but the function $f$ cannot be used for that. For the last steps we will use the initial bound (1):

\begin{equation*}
LCI\_ORS(p_i,D) - LCI\_ORS(p_j,D) \geq \ln \left(1+\frac{1}{c}\right) - 2 \frac{ \frac{1}{c(c+1)}} {\sqrt{\frac{1}{c+1}}  + \sqrt{\frac{1}{c}} }
\end{equation*}

If $c=1$

\begin{equation*}
LCI\_ORS(p_i,D) - LCI\_ORS(p_j,D) = \ln 2 - \frac{1}{1 + \sqrt{\frac{1}{2}} } \geq 0
\end{equation*}

If $c=2$

\begin{equation*}
LCI\_ORS(p_i,D) - LCI\_ORS(p_j,D) = \ln \frac{3}{2} - \frac{\frac{1}{3}}{ \sqrt{ \frac{1}{2}} + \sqrt{\frac{1}{3}} } \geq 0
\end{equation*}

If $c=3$

\begin{equation*}
LCI\_ORS(p_i,D) - LCI\_ORS(p_j,D) = \ln \frac{4}{3} - \frac{\frac{1}{6}} {\frac{1}{2} + \sqrt{\frac{1}{3}} } \geq 0
\end{equation*}

Eventually, for all $c \geq 1$ we have $LCI\_ORS(p_i,D) > LCI\_ORS(p_j,D)$. Gathering the cases $d \leq c+1$ and $d \geq c+2 $, we have $LCI\_ORS(p_i,D) > LCI\_ORS(p_j,D)$ for all $a,b,c,d \in \mathbb{N}^* $ with $d \geq 2$.

\textbf{Demonstrate $LCI\_GR(p_i,D) > LCI\_GR(p_j,D)$ }

Lower confidence intervals of $GR$ of $p_i$ and $p_j$ are given:

\begin{equation*}
LCI\_GR(p_i,D) = \exp \left( ln(\frac{a(c+d)}{c(a+b)}) - 1.96 \sqrt{\frac{1}{a} - \frac{1}{a+b} + \frac{1}{c} - \frac{1}{c+d}} \right)
\end{equation*}

\begin{equation*}
LCI\_GR(p_j,D) = \exp \left( ln(\frac{a(c+d)}{(c+1)(a+b)}) - 1.96 \sqrt{\frac{1}{a} - \frac{1}{a+b} + \frac{1}{c+1} - \frac{1}{c+d}} \right)
\end{equation*}

We want to approve $LCI\_GR(p_i,D) > LCI\_GR(p_j,D)$. Similar with proof of lower confidence interval of $ORS$, we want to demonstrate that $LCI\_GR(p_i,D) - LCI\_GR(p_j,D) > 0$, we rewrite this inequality as follow: 

\begin{equation*}
g_4 = ln(\frac{c+1}{c})  - 1.96  \frac{ \frac{1}{c(c+1)}}{\sqrt{\frac{1}{a} - \frac{1}{a+b} + \frac{1}{c} - \frac{1}{c+d}} + \sqrt{\frac{1}{a} - \frac{1}{a+b} + \frac{1}{c+1} - \frac{1}{c+d}}}  >0
\end{equation*}

We set $\alpha = \frac{1}{a} - \frac{1}{a+b} > 0 $ and we have:

\begin{equation*}
g_4 > g_2 = \ln \left(1+\frac{1}{c}\right) - 1.96  \frac{ \frac{1}{c(c+1)}} {\sqrt{\frac{1}{c} - \frac{1}{c+d}} + \sqrt{\frac{1}{c+1} - \frac{1}{c+d}}}
\end{equation*}

We try to delete $d$ in this lower bound:

\begin{equation*}
\frac{1}{c} - \frac{1}{c+d} \geq \frac{1}{c} - \frac{1}{c+1} = \frac{1}{c(c+1)}
\end{equation*}

\begin{equation*}
\frac{1}{c+1} - \frac{1}{c+d} \geq \frac{1}{c+1} - \frac{1}{c+1} = 0
\end{equation*}

The second inequality involves problems to lower-bound $g_2$ (this lower bound will not be positive). So, in the next part we assume $d \geq 2$. Thus:

\begin{equation*}
\frac{1}{c} - \frac{1}{c+d} \geq \frac{1}{c} - \frac{1}{c+2} = \frac{2}{c(c+2)}
\end{equation*}

\begin{equation*}
\frac{1}{c+1} - \frac{1}{c+d} \geq \frac{1}{c+1} - \frac{1}{c+2} = \frac{1}{(c+1)(c+2)}
\end{equation*}

We get:

\begin{equation*}
g_2 \geq g_1 = \ln \left(1+\frac{1}{c}\right) - 1.96 \frac{ \frac{1}{c(c+1)}} { \sqrt{\frac{2}{c(c+2)}} + \sqrt{\frac{1}{(c+1)(c+2)}}}
\end{equation*}

We can simplify a little:

\begin{equation*}
g_1 = \ln \left(1+\frac{1}{c}\right) - 1.96 \sqrt{\frac{c+2}{c(c+1)}} \textrm{.}\frac{1}{\sqrt{c} + \sqrt{2(c+2)}}
\end{equation*}

We have to show that this function is positive for all $c \geq 1$. However, this is not the case for $c=1$ and $c=2$ but we can show that it is true for all $c \geq 3$. Hence, the issue is that $g_1$ is not directly easy to analyze, so we have to provide a easier lower bound but it implies some singular cases to treat.

\begin{equation*}
g_1 \geq \ln \left(1+\frac{1}{c}\right) - 1.96 \frac{1}{\sqrt{c}} \textrm{.} \sqrt{\frac{c+2}{c+1}} \textrm{.} \frac{1}{\sqrt{c}+\sqrt{2c}}
\end{equation*}

\begin{equation*}
\geq \ln \left(1+\frac{1}{c}\right) - \frac{\beta}{c} \textrm{.} \sqrt{1+\frac{1}{c+1}}
\end{equation*}

\begin{equation*}
\geq \ln \left(1+\frac{1}{c}\right) - \frac{\beta}{c} \textrm{.} \sqrt{1+\frac{1}{c}}
\end{equation*}

Where $\beta = \frac{1.96}{1+\sqrt{2}} \simeq 0.812$. Let us use $f(c)$ as new lower bound. We can calculate:

\begin{equation*}
f'(c) = -\frac{1}{c(c+1)} \left(1-\beta \sqrt{1+\frac{1}{c}} \left(1+\frac{3}{2c}\right) \right) = - \frac{1}{c(c+1)} \textrm{.} g(c)
\end{equation*}

We want to show that the lower bound function $f$ is positive. Actually, we are going to show that this function is decreasing beyond some point. First we can notice that $g$ is increasing:

\begin{equation*}
U : x \mapsto 1 - \beta \sqrt{1+x} \left(1+\frac{3}{2}x\right) \textrm{ and } V : x \mapsto \frac{1}{x} \textrm{ are clearly decreasing on } \mathbb{R}^+
\end{equation*}

So $g= U \circ V$ is increasing on $\mathbb{R}^+$. Moreover $g(9)>0$, then for all $c \geq 9$, $g(c) \geq 0$. It implies that for all $c \geq 9$, $f'(c) \leq 0$. Nonetheless, we notice that $f(9) \geq 0$ and $f(c) \mapsto 0$ when $c \mapsto \infty$. As $f$ is decreasing for $c \geq 9$, it means that $f$ is positive for $c \geq 9$. We sum up (for $d \geq 2, c \geq 9$):

\begin{equation*}
LCI\_GR(p_i,D) - LCI\_GR(p_j,D) \geq g_4 > g_1 \geq f(c) \geq 0
\end{equation*}

The cases $(c=1,d > 1)$ and $(c=2, d > 1)$ need to be treated. We will use the $g_2$ function, we recall:

\begin{equation*}
g_2 = \ln \left(1+\frac{1}{c}\right) - 1.96  \frac{ \frac{1}{c(c+1)}} {\sqrt{\frac{1}{c} - \frac{1}{c+d}} + \sqrt{\frac{1}{c+1} - \frac{1}{c+d}}}
\end{equation*}

Let us set $c=2$, then:

\begin{equation*}
g_2 = \ln \left(\frac{3}{2}\right) - \frac{1.96}{6} \frac{1} {\sqrt{\frac{1}{2} - \frac{1}{2+d}} + \sqrt{\frac{1}{3} - \frac{1}{2+d}}}
\end{equation*}

\begin{equation*}
\textrm{The function } P:x \mapsto \ln \left(\frac{3}{2}\right) - \frac{1.96}{6} \textrm{.}x \textrm{ and } Q:x \mapsto \frac{1} {\sqrt{\frac{1}{2} + x} + \sqrt{\frac{1}{3} + x}}
\end{equation*}

are decreasing and $R:x \mapsto -\frac{1}{d+x}$ is increasing, so that $d \mapsto g_2 = P \circ Q \circ R$ is increasing. Unfortunately $g_2$ with $c=2,d=2$ is negative but with $c=2,d=3$ is positive. It means that the inequality is true in the case $(c=2,d=3)$ because $g_2$ is increasing, but not for the case $(c=2,d=2)$. In the same way, we can show that the inequality is true for the case $(c=1,d \geq 4)$ but not for the remaining cases: $(c=1,d=2)$ and $(c=1,d=3)$.

The inequality $LCI\_GR(p_i,D) - LCI\_GR(p_j,D) > 0$ is true except the cases $(c=1,d=2)$, $(c=1,d=3)$ and $(c=2,d=2)$. However, we have $c+d=|D_2|$, is a large integer. Thus these remaining cases cannot be happened in practice.

\end{document}